\documentclass[conference]{IEEEtran}
\IEEEoverridecommandlockouts
\ifCLASSINFOpdf
\else
\fi
\usepackage{amsmath,amsfonts,amssymb,amsthm,bm}

\def\Tableref#1{Table~\ref{#1}}

\def\Figref#1{Figure~\ref{#1}}

\def\Secref#1{Section~\ref{#1}}

\def\eqref#1{equation~(\ref{#1})}
\def\twoeqrefs#1#2{equations (\ref{#1}) and (\ref{#2})}
\def\Eqref#1{Equation~(\ref{#1})}  
\def\plaineqref#1{(\ref{#1})}  

\def\Algref#1{Algorithm~\ref{#1}}

\def\1{\bm{1}}

\def\rh{{\textnormal{h}}}

\def\rx{{\textnormal{x}}}
\def\ry{{\textnormal{y}}}
\def\rz{{\textnormal{z}}}

\def\rvh{{\mathbf{h}}}

\def\rvx{{\mathbf{x}}}

\def\rvz{{\mathbf{z}}}

\def\rmH{{\mathbf{H}}}

\def\rmX{{\mathbf{X}}}

\def\valpha{{\bm{\alpha}}}
\def\vbeta{{\bm{\beta}}}
\def\vmu{{\bm{\mu}}}

\def\vpi{{\bm{\pi}}}
\def\vlambda{{\bm{\lambda}}}
\def\vsigma{{\bm{\sigma}}}

\def\vx{{\bm{x}}}
\def\vy{{\bm{y}}}

\def\evalpha{{\alpha}}

\def\evpi{{\pi}}

\def\mA{{\bm{A}}}

\def\mG{{\bm{G}}}

\def\mI{{\bm{I}}}

\def\mX{{\bm{X}}}

\DeclareMathAlphabet{\mathsfit}{\encodingdefault}{\sfdefault}{m}{sl}
\SetMathAlphabet{\mathsfit}{bold}{\encodingdefault}{\sfdefault}{bx}{n}
\newcommand{\tens}[1]{\bm{\mathsfit{#1}}}

\def\tX{{\tens{X}}}

\def\gG{{\mathcal{G}}}

\def\sP{{\mathbb{P}}}

\def\sY{{\mathbb{Y}}}

\def\emA{{A}}

\def\emG{{G}}

\newcommand{\E}{\mathbb{E}}

\newcommand{\R}{\mathbb{R}}

\newcommand{\reg}{\lambda}

\newcommand{\KL}{D_{\mathrm{KL}}}

\newcommand{\normlone}{\ell^1}
\newcommand{\normltwo}{\ell^2}

\DeclareMathOperator*{\argmax}{arg\,max}
\DeclareMathOperator*{\argmin}{arg\,min}

\newcommand{\bcdot}{\boldsymbol{\cdot}}
\newcommand{\transp}{^\intercal} 

\usepackage[bookmarks=false]{hyperref}
\usepackage{url}
\usepackage{amsfonts}
\usepackage{amssymb}
\usepackage{algorithm}
\usepackage{algorithmicx}
\usepackage{algpseudocode}
\usepackage{mathtools}
\usepackage{amsthm}
\usepackage{mathabx}
\usepackage[bordercolor=white,backgroundcolor=gray!30,linecolor=black,colorinlistoftodos]{todonotes}
\usepackage{nameref}
\usepackage{pgfplots}
\pgfplotsset{compat=newest}
\usepackage{caption}

\newcommand{\thetaprev}{\theta^\text{-}}
\newcommand{\Indyk}{\1_{y_i=k}}
\newcommand{\Indbeta}{\1_{\beta_{k,m} > 0}}
\newcommand{\tb}{\textbf}
\newcommand{\tu}{\underline}
\newtheorem{proposition}{Proposition}

\renewcommand{\IEEEQED}{\IEEEQEDopen} 

\begin{document}

\begin{titlepage}
\copyright~2019 IEEE. Personal use of this material is permitted. Permission from IEEE must be obtained for all other uses, in any current or future media, including reprinting/republishing this material for advertising or promotional purposes, creating new collective works, for resale or redistribution to servers or lists, or reuse of any copyrighted component of this work in other works.\\ 

%

Accepted at the 2019 International Joint Conference on Neural Networks (IJCNN). 
\end{titlepage}

%

\title{SpaMHMM: Sparse Mixture of Hidden Markov Models for Graph Connected Entities\\
\thanks{This work was partially financed by the ERDF – European Regional Development Fund through the Operational Programme for Competitiveness and Internationalisation - COMPETE 2020 Programme and by National Funds through the Portuguese funding agency, FCT - Funda\c{c}\~{a}o para a Ci\^{e}ncia e a Tecnologia within project ``POCI-01-0145-FEDER-028857'' and also by Funda\c{c}\~{a}o para a Ci\^{e}ncia e a Tecnologia within Ph.D. grant number SFRH/BD/129600/2017.}
}

\author{\IEEEauthorblockN{Diogo Pernes}
\IEEEauthorblockA{\textit{INESC TEC} \\
\textit{University of Porto}\\
Porto, Portugal \\
dpc@inesctec.pt}
\and
\IEEEauthorblockN{Jaime S. Cardoso}
\IEEEauthorblockA{\textit{INESC TEC} \\
\textit{University of Porto}\\
Porto, Portugal \\
jaime.cardoso@inesctec.pt}}


%


\maketitle

\begin{abstract}
We propose a framework to model the distribution of sequential data coming from a set of entities connected in a graph with a known topology. The method is based on a mixture of shared hidden Markov models (HMMs), which are jointly trained in order to exploit the knowledge of the graph structure and in such a way that the obtained mixtures tend to be sparse. Experiments in different application domains demonstrate the effectiveness and versatility of the method.
\end{abstract}

%

\section{Introduction and state of the art}
Hidden Markov models (HMMs) are a ubiquitous tool for modelling sequential data. They started by being applied to speech recognition systems and from there they have spread to almost any application one can think of, encompassing computational molecular biology, data compression, and computer vision. In the emerging field of cognitive radars~(\cite{Greco2018}), for the task of opportunistic usage of the spectrum, HMMs have been recently used to model the occupancy of the channels by primary users~\cite{Stinco2016}. 

When the expressiveness of an HMM is not enough, mixtures of HMMs have been adopted. Roughly speaking, mixtures of HMMs can be interpreted as the result of the combination of a set of independent standard HMMs which are observed through a memoryless transformation~\cite{Couvreur1996,Dias2010,NIPS2014_5518,Helske2016MixtureHM}.

In many real-life settings one does not have a single data stream but an  arbitrary number of network connected entities that share and interact in the same medium and generate data streams in real-time. The streams produced by each of these entities form a set of time series with both intra and inter relations between them. In neuroimaging studies, the brain can be regarded as a network: a connected system where nodes, or units, represent different specialized regions and links, or connections, represent communication pathways. From a functional perspective, communication is coded by temporal dependence between the activities of different brain areas~\cite{DeVicoFallani20130521}. Also team sports intrinsically involve fast, complex and interdependent events among a set of entities (the players), which interact as a team~\cite{Tora2017, Theagarajan2018}. Thus, in all these scenarios the behavior of each individual entity is better understood if its context information (i.e. the behavior of the neighboring instances) is leveraged.

The extraction of knowledge from these streams to support the decision-making process is still challenging and the adaptation of HMM to this scenario is immature at best.
\cite{FERLES2013133} proposed a hybrid approach combining the Self-Organizing Map (SOM) and the HMM with applications in clustering, dimensionality reduction and visualization of large-scale sequence spaces. Note that the model at each node is limited to a simple HMM.
Wireless local area networks have also been modeled with Markov-based approaches. For instance, \cite{Anisa2014} use HMMs for outlier detection in 802.11 wireless access points. However, the typical approaches include a common HMM model for all nodes (with strong limited flexibility) and a HMM model per node, independent of the others (not exploring the dependencies between nodes).
\cite{Bolton2018} built a sparse coupled hidden Markov model (SCHMM) framework to parameterize the temporal evolution of data acquired with functional magnetic resonance imaging (fMRI). The coupling is captured in the transition matrix, which is assumed to be a function of the activity levels of all the streams; the model per node is still restricted to a simple HMM.

In general, in networked data streams, the stream observed in each sensor is often modeled by HMMs but the intercorrelations between sensors are seldom explored. The proper modeling of the intercorrelations has the potential to improve the learning process, acting as a regularizer in the learning process.
In this work, we aim to tackle this void, by proposing as observation model at each node a sparse mixture of HMMs, where the dependencies between nodes are used to promote the sharing of HMM components between similar nodes.

\section{Proposed model}
The proposed model finds intersections with distributed sparse representations.  

\subsection{Overview}
Sparse representation/coding expresses a signal/model $f$, defined over some independent variable $x$, as a linear combination of a few atoms from a prespecified and overcomplete dictionary of size $M$:
\begin{equation}
\label{sparse_coding}
f(x)=\sum_{m=1}^M s_m \phi_m(x),
\end{equation}
where $\phi_m(x)$ are the atoms and only a few of the scalars $s_m$ are non-zero, providing a sparse representation of $f(x)$.

Distributed sparse representation \cite{Baron} is an extension of the standard version that considers networks with $K$ nodes.
At each node, the signal sensed at the same node has its sparsity property because of its intracorrelation, while, for networks with multiple nodes, signals received at different nodes also exhibit strong intercorrelation.
The intra- and inter-correlations lead to a joint sparse model. An interesting scenario in distributed sparse representation is when all signals/models share the common support but with different non-zero coefficients.

Inspired by the formulation of \eqref{sparse_coding}, we propose to model the generative distribution of the data coming from each of the $K$ nodes of a network as a sparse mixture obtained from a dictionary of generative distributions. Specifically, we shall model the distribution for each node as a sparse mixture over a `large' shared dictionary of HMMs, where each HMM corresponds to an individual atom from the dictionary.
The field knowledge about the similarities between nodes is summarized in an affinity matrix. The objective function of the learning process promotes reusing HMM atoms between similar nodes.
We now formalize these ideas.

\subsection{Model formulation}
\subsubsection{Definition}
\label{sec:definition}
Assume we have a set of nodes $\sY=\{1, ..., K\}$ connected by an undirected weighted graph $\gG$, expressed by a symmetric matrix $\mG \in \R^{K \times K}$. These nodes thus form a network, in which the weights are assumed to represent degrees of affinity between each pair of nodes (i.e. the greater the edge weight, the more the respective nodes \textit{like} to agree). The nodes $\ry$ in the graph produce $D$-dimensional sequences $\rmX = \left(\rvx^{(1)}, ...,\rvx^{(T)} \right)$, $\rvx^{(t)} \in \R^D$, whose conditional distribution we shall model using a mixture of HMMs:
\begin{equation}
\label{hmm_mix}
p(\rmX | \ry) = \sum_{\rz} p(\rz|\ry) p(\rmX | \rz),
\end{equation}
where $\rz \in \{1, ..., M\}$ is a latent random variable, being $M$ the size of the mixture. This is a particular realization of \eqref{sparse_coding} where $f$ is the probability density function $p(\rmX | \ry)$ and the coefficients $s_m$ correspond to the probabilities $p(\rz = m | \ry)$. Here, $p(\rmX | \rz)$ is the marginal distribution of observations of a standard first-order homogeneous HMM:
\begin{equation}
\label{hmm}
p(\rmX | \rz) = \sum_{\rvh} p(\rh^{(0)}|\rz) \prod_t p(\rh^{(t)}|\rh^{(t-1)},\rz) p(\rvx^{(t)}|\rh^{(t)},\rz),
\end{equation}
where $\rvh = \left(\rh^{(0)}, ...,\rh^{(T)} \right)$, $\rh^{(t)} \in \{1, ..., S\}$, is the sequence of hidden states of the HMM, being $S$ the number of hidden states. Note that the factorization in \eqref{hmm_mix} imposes conditional independence between the sequence $\rmX$ and the node $\ry$, given the latent variable $\rz$. This is a key assumption of this model, since this way the distributions for the observations in the nodes in $\sY$ share the same dictionary of HMMs, promoting parameter sharing among the $K$ mixtures.

\subsubsection{Inference}
\label{sec:inference}
Given an observed sequence $\mX$ and its corresponding node $y \in \sY$, the inference problem here consists in finding the likelihood $p(\rmX=\mX | \ry=y)$ (from now on, abbreviated as $p(\mX|y)$) as defined by \twoeqrefs{hmm_mix}{hmm}. The marginals $p(\mX|\rz)$ of each HMM in the mixture may be computed efficiently, in $O(S^2T)$ time, using the Forward algorithm~\cite{Rabiner1986}. Then, $p(\mX|y)$ is obtained by applying \eqref{hmm_mix}, so inference in the overall model is done in at most $O(MS^2T)$ time. As we shall see, however, the mixtures we get after learning will often be sparse (see \Secref{sec:learning}), leading to an even smaller time complexity.

\subsubsection{Learning}
\label{sec:learning}
Given an i.i.d. dataset consisting of $N$ tuples $(\mX_i, y_i)$ of sequences of observations $\mX_i = \left(\vx_i^{(1)}, ...,\vx_i^{(T_i)} \right)$ and their respective nodes $y_i \in \sY$, the model defined by \twoeqrefs{hmm_mix}{hmm} may be easily trained using the Expectation-Maximization (EM) algorithm \cite{Dempster1977}, (locally) maximizing the usual log-likelihood objective:
\begin{equation}
\label{log_likelihood}
J(\theta) = \sum_{i=1}^N \log p(\mX_i | y_i, \theta),
\end{equation}
where $\theta$ represents all model parameters, namely:
\begin{enumerate}
\item the $M$-dimensional mixture coefficients, $\valpha_k \coloneqq \left(p(\rz=1|\ry=k), ..., p(\rz=M|\ry=k)\right)$, for $k = 1,...,K$;
\item the $S$-dimensional initial state probabilities, $\vpi_m \coloneqq \left(p(\rh^{(0)}=1 | \rz=m),...,p(\rh^{(0)}=S | \rz=m)\right)$, for $m = 1,...,M$;
\item the $S \times S$ state transition matrices, $\mA^m$, where $\emA^m_{s,u} \coloneqq p(\rh^{(t)}=u | \rh^{(t-1)}=s, \rz=m)$, for $s,u = 1,...,S$ and $m = 1,...,M$;
\item the emission probability means, $\vmu_{m,s} \in \mathbb{R}^D$, for $m = 1,...,M$ and $s = 1,...,S$;
\item the emission probability diagonal covariance matrices, $\mI \vsigma^2_{m,s}$, where $\vsigma^2_{m,s} \in \mathbb{R}^{D}_{+}$, for $m = 1,...,M$ and $s = 1,...,S$.
\end{enumerate}

Here, we are assuming that the emission probabilities $p(\rx^{(t)}|\rh^{(t)},\rz)$ are Gaussian with diagonal covariances. This introduces almost no loss of generality, since the extension of this work to discrete observations or other types of continuous emission distributions is straightforward.

The procedure to maximize objective~\plaineqref{log_likelihood} using EM is described in \Algref{alg:mhmm}. The update formulas follow from the standard EM procedure and can be obtained by viewing this model as a Bayesian network or by following the derivation detailed in \Secref{sec:proof_em_noreg}. However, the objective \plaineqref{log_likelihood} does not take advantage of the known structure of $\gG$. In order to exploit this information, we introduce a regularization term, maximizing the following objective instead:
\begin{align}
\label{objective}
J_r(\theta) &= \frac{1}{N}\sum_{i=1}^N \log p(\mX_i | y_i, \theta) \nonumber\\ 
&\mathbin{\hphantom{=}}{}+ \frac{\reg}{2} \sum_{\substack{j,k=1,\\k\neq j}}^{K} \emG_{j,k} \E_{\rz \sim p(\rz | \ry=j, \theta)} [ p(\rz | \ry=k, \theta) ] \nonumber\\
&= \frac{1}{N}\sum_{i=1}^N \log p(\mX_i | y_i, \theta) + \frac{\reg}{2} \sum_{\substack{j,k=1,\\k\neq j}}^{K} \emG_{j,k} \valpha_j \transp \valpha_k,
\end{align}
where $\reg \geq 0$ controls the relative weight of the two terms in the objective. Note that this regularization term favors nodes connected by edges with large positive weights to have similar mixture coefficients and thus share mixture components. On the other hand, nodes connected by edges with large negative weights will tend to have orthogonal mixture coefficients, being described by disjoint sets of components. These observations agree with our prior assumption that the edge weights express degrees of similarity between each pair of nodes. Proposition \ref{prop_expectations} formalizes these statements and enlightens interesting properties about the expectations $\E_{\rz \sim p(\rz | \ry=j, \theta)} [ p(\rz | \ry=k, \theta)]$.
\begin{proposition}
\label{prop_expectations}
Let $\sP_M$ be the set of all $M$-nomial probability distributions and $M>1$. We have:
\begin{enumerate}
\item $\min_{p,q \in \sP_M} \E_{\rz\sim p} [ q(\rz) ] = 0$; \label{prop_min}
\item $\argmin_{p,q \in \sP_M} \E_{\rz\sim p} [ q(\rz) ] = \{p,q \in \sP_M \mid \forall \, m \in \{1,...,M\}: p(\rz = m)q(\rz = m)=0\}$; \label{prop_argmin}
\item $\max_{p,q \in \sP_M} \E_{\rz\sim p} [ q(\rz) ] = 1$; \label{prop_max}
\item $\argmax_{p,q \in \sP_M} \E_{\rz\sim p} [ q(\rz) ] = \{p,q \in \sP_M \mid \exists \, m \in \{1,...,M\} : \, p(\rz = m)=q(\rz = m)=1\}$ \label{prop_argmax}.
\end{enumerate}
\end{proposition}
\begin{IEEEproof}
By the definition of expectation,
\begin{equation}
\label{prop_expdef}
\E_{\rz\sim p} [ q(\rz) ] = \sum_{m=1}^M p(\rz=m) q(\rz=m).
\end{equation}
Statements \ref{prop_min} and \ref{prop_argmin} follow immediately from the fact that every term in the right-hand side of \plaineqref{prop_expdef} is non-negative and $M>1$. For the remaining, we rewrite \plaineqref{prop_expdef} as the dot product of two $M$-dimensional vectors $\valpha_p$ and $\valpha_q$, representing the two distributions $p$ and $q$, respectively, and we use the following linear algebra inequalities to build an upper bound for this expectation:
\begin{equation}
\label{prop_ineq}
\E_{\rz\sim p} [ q(\rz) ] = \valpha_p \transp \valpha_q \leq || \valpha_p ||_2 || \valpha_q ||_2 \leq || \valpha_p ||_1 || \valpha_q ||_1 =1,
\end{equation}
where $||\bcdot||_1$ and $||\bcdot||_2$ are the $\normlone$ and $\normltwo$ norms, respectively. Clearly, the equality $\E_{\rz\sim p} [ q(\rz) ] = 1$ holds if $p$ and $q$ are chosen from the set defined in statement \ref{prop_argmax}, where the distributions $p$ and $q$ are the same and they are non-zero for a single assignment of $\rz$. This proves statement \ref{prop_max}. Now, to prove statement \ref{prop_argmax}, it suffices to show that there are no other maximizers. The first inequality in \plaineqref{prop_ineq} is transformed into an equality if and only if $\valpha_p = \valpha_q$, which means $p \equiv q$. The second inequality becomes an equality when the $\normlone$ and $\normltwo$ norms of the vectors coincide, which happens if and only if the vectors have only one non-zero component, concluding the proof.
\end{IEEEproof}
Specifically, given two distinct nodes $j, k \in \sY$ , if $\emG_{j,k} > 0$, the regularization term for these nodes is maximum (and equal to $\emG_{j,k}$) when the mixtures for these two nodes are the same and have one single active component (i.e. one mixture component whose coefficient is non-zero). On the contrary, if $\emG_{j,k} < 0$, the term is maximized (and equal to zero) when the mixtures for the two nodes do not share any active components. In both cases, though, we conclude from Proposition \ref{prop_expectations} that we are favoring sparse mixtures. We see sparsity as an important feature since it allows the size $M$ of the dictionary of models to be large and therefore expressive without compromising our rational that the observations in a given node are well modeled by a mixture of only a few HMMs. This way, some components will specialize on describing the behavior of some nodes, while others will specialize on different nodes. Moreover, sparse mixtures yield faster inference, more interpretable models and (possibly) less overfitting.



By setting $\reg=0$, we clearly get the initial objective \plaineqref{log_likelihood}, where inter-node correlations are modeled only via parameter sharing. As $\reg \to \infty$, two interesting scenarios may be anticipated. If $\emG_{j,k} > 0, \, \forall j,k \in \sY,$ all nodes will tend do share the same single mixture component, i.e. we would be learning one single HMM to describe the whole network. If $\emG_{j,k} < 0, \, \forall j,k \in \sY,$ and $M \geq K$, each node would tend to learn its own HMM model independently from all the others. Again, in both scenarios, the obtained mixtures are sparse.

The objective function~\plaineqref{objective} can still be maximized via EM (see details in \Secref{sec:proof_em_reg}). However, the introduction of the regularization term in the objective makes it impossible to find a closed form solution for the update formula of the mixture coefficients. Thus, in the M-step, we need to resort to gradient ascent to update these parameters. In order to ensure that the gradient ascent iterative steps lead to admissible solutions, we adopt the following reparameterization from \cite{Yang2018}:
\begin{equation}
\label{normalization}
\alpha_{k,m} = \frac{\sigma \left(\beta_{k,m} \right)^2}{\sum_{l=1}^M \sigma \left( \beta_{k,l} \right)^2}, 
\end{equation}
for $k = 1, ..., K$ and $m = 1, ..., M$, and where $\sigma(\bcdot)$ is the rectifier linear (ReLU) function. This reparameterization clearly resembles the softmax function, but, contrarily to that one, admits sparse outputs. The squared terms in \eqref{normalization} aim only to make the optimization more stable. The optimization steps for the objective \plaineqref{objective} using this reparameterization are described in \Algref{alg:spamhmm}.

\begin{algorithm}
\caption{EM algorithm for the mixture without regularization (MHMM).}
\label{alg:mhmm}
\begin{algorithmic}
    \State Inputs: The training set, consisting of $N$ tuples $(\mX_i,y_i)$, a set of initial parameters $\theta^{(0)}$ and the number of training iterations $\mathcal{I}$.
    \For{$j = 1, ..., \mathcal{I}$}
        \State Sufficient statistics:
        \begin{enumerate}
            \item $n_k \coloneqq \sum_i \Indyk$, where $\1_\mathrm{(\bcdot)}$ is the indicator function, for $k=1,...,K$.
            \item Obtain the mixture posteriors $\eta_{i,m} \coloneqq p(\rz=m|\mX_i, y_i,\theta^{(j-1)})$, for $i=1,...,N$ and $m=1,...,M$, by computing  $\tilde{\eta}_{i,m} \coloneqq p(\mX_i|\rz=m, \theta^{(j-1)})p(\rz=m|y_i, \theta^{(j-1)})$ and normalizing it.
            \item Obtain the state posteriors $\gamma_{i,m,s}(t) \coloneqq p(\rh^{(t)}=s | \rz=m, \mX_i, \theta^{(j-1)})$ and $\xi_{i,m,s,u}(t) \coloneqq p(\rh^{(t-1)}=s, \rh^{(t)}=u | \rz=m, \mX_i, \theta^{(j-1)})$, for $i=1,...,N$, $m=1,...,M$ and $s,u=1,...,S$, as done in the Baum-Welch algorithm \cite{Baum1972}.
        \end{enumerate}
        \State M-step:
        \begin{enumerate}
            \item $\evalpha_{k,m} = \frac{\sum_i \eta_{i,m} \Indyk}{n_k}$, for $k=1,...,K$ and $m=1,...,M$, obtaining $\valpha_k$.
            \item $\evpi_{m,s} = \frac{\sum_i \eta_{i,m} \gamma_{i,m,s}(0)}{\sum_i \eta_{i,m}}$, for $m=1,...,M$ and $s=1,...,S$, obtaining $\vpi_{m}$. 
            \item $\emA^m_{s,u} = \frac{\sum_i \eta_{i,m} \sum_{t=1}^{T_i} \xi_{i,m,s,u}(t)}{\sum_i \eta_{i,m} \sum_{t=0}^{T_i-1} \gamma_{i,m,s}(t)}$, for $m=1,...,M$ and $s,u=1,...,S$, obtaining $\mA^m$.
            \item $\vmu_{m,s} = \frac{\sum_i \eta_{i,m} \sum_{t=1}^{T_i} \gamma_{i,m,s}(t) \vx_i^{(t)}} {\sum_i \eta_{i,m} \sum_{t=1}^{T_i} \gamma_{i,m,s}(t)}$, for $m=1,...,M$ and $s=1,...,S$.
            \item $\vsigma^2_{m,s} = \frac{\sum_i \eta_{i,m} \sum_{t=1}^{T_i} \gamma_{i,m,s}(t) \left(\vx_i^{(t)} - \boldsymbol{\mu}^m_s\right)^2} {\sum_i \eta_{i,m} \sum_{t=1}^{T_i} \gamma_{i,m,s}(t)}$, for $m=1,...,M$ and $s=1,...,S$.
          \item $\theta^{(j)} = \bigcup_{k,m,s} \left\lbrace \boldsymbol{\alpha}_k, \vpi_{m}, \mA^m, \vmu_{m,s}, \vsigma^2_{m,s} \right\rbrace$.
        \end{enumerate}
    \EndFor  
\end{algorithmic}
\end{algorithm}

\begin{algorithm}
\caption{EM algorithm for the mixture with regularization (SpaMHMM).}
\label{alg:spamhmm}
\begin{algorithmic}
\State Inputs: The training set, consisting of $N$ tuples $(\mX_i,y_i)$, the matrix $\mG$ describing the graph $\gG$, the regularization hyperparameter $\reg$, a set of initial parameters $\theta^{(0)}$, the number of training iterations $\mathcal{I}$, the number of gradient ascent iterations $\mathcal{J}$ to perform on each M-step, the learning rate $\rho$ for the gradient ascent.
\For{$j=1,...,\mathcal{I}$}
    \State Sufficient statistics: same as in \Algref{alg:mhmm}.
    \State M-step:
	\For{$l=1,...,\mathcal{J}$}
        \begin{enumerate}
            \item $\psi_{k,m} \coloneqq \frac{1}{N}\sum_i (\eta_{i,m} - \alpha_{k,m}) \Indyk$, for $k=1,...,K$ and $m=1,...,M$.
            \item $\omega_{k,m} \coloneqq \alpha_{k,m} \sum_{j \neq k} \emG_{j,k}\left(\alpha_{j,m} - \valpha_j \transp \valpha_k \right)$, for $k=1,...,K$ and $m=1,...,M$.
            \item $\delta_{k,m} \coloneqq \Indbeta \frac{2\sigma'(\beta_{k,m})}{\sigma(\beta_{k,m})}\left(\psi_{k,m} + \reg \omega_{k,m}\right)$, where $\sigma'(\cdot)$ is the derivative of $\sigma(\cdot)$, for $k=1,...,K$ and $m=1,...,M$. 
            \item $\beta_{k,m} \leftarrow \beta_{k,m} + \rho \delta_{k,m}$, for $k=1,...,K$ and $m=1,...,M$.
            \item Use \eqref{normalization} to obtain $\alpha_{k,m}$, for $k=1,...,K$ and $m=1,...,M$.
        \end{enumerate}
    \EndFor
    \State Do steps 2) -- 6) in the M-step of \Algref{alg:mhmm}. 
    \EndFor
\end{algorithmic}
\end{algorithm}

\section{Experimental Evaluation}
\label{sec:experiments}
The model was developed on top of the library hmmlearn~\cite{hmmlearn} for Python, which implements inference and unsupervised learning for the standard HMM using a wide variety of emission distributions. Both learning and inference use the hmmlearn API, with the appropriate adjustments for our models. For reproducibility purposes, we make our source code, pre-trained models and the datasets publicly available\footnote{\url{https://github.com/dpernes/spamhmm}}.

We evaluate four different models in our experiments: a model consisting of a single HMM (denoted as 1-HMM) trained on sequences from all graph nodes; a model consisting of $K$ HMMs trained independently (denoted as K-HMM), one for each graph node; a mixture of HMMs (denoted as MHMM) as defined in this work (\twoeqrefs{hmm_mix}{hmm}), trained to maximize the usual log-likelihood objective~\plaineqref{log_likelihood}; a mixture of HMMs (denoted as SpaMHMM) as the previous one, trained to maximize our regularized objective~\plaineqref{objective}.
Models 1-HMM, K-HMM and MHMM will be our baselines. We shall compare the performance of these models with that of SpaMHMM and, for the case of MHMM, we shall also verify if SpaMHMM actually produces sparser mixtures in general, as argued in \Secref{sec:learning}. In order to ensure a fair comparison, we train models with approximately the same number of possible state transitions. Hence, given an MHMM or SpaMHMM with $M$ mixture components and $S$ states per component, we train a 1-HMM with $\approx S\sqrt[]{M}$ states and a K-HMM with $\approx S\sqrt[]{M/K}$ states per HMM. We initialize the mixture coefficients in MHMM and SpaMHMM randomly, while the state transition matrices and the initial state probabilities are initialized uniformly. Means are initialized using $k$-means, with $k$ equal to the number of hidden states in the HMM, and covariances are initialized with the diagonal of the training data covariance. Models 1-HMM and K-HMM are trained using the Baum-Welch algorithm, MHMM is trained using \Algref{alg:mhmm} and SpaMHMM is trained using \Algref{alg:spamhmm}. However, we opted to use Adam \cite{Kingma2014} instead of \textit{vanilla} gradient ascent in the inner loop of \Algref{alg:spamhmm}, since its per-parameter learning rate proved to be beneficial for faster convergence. 

\subsection{Anomaly detection in Wi-Fi networks}
\label{sec:wi_fi}
A typical Wi-Fi network infrastructure is constituted by $K$ access points (APs) distributed in a given space. The network users may alternate between these APs seamlessly, usually connecting to the closest one. There is a wide variety of anomalies that may happen during the operation of such network and their automatic detection is, therefore, of great importance for future mitigation plans. Some anomalous behaviors are: overloaded APs, failed or crashed APs, persistent radio frequency interference between adjacent APs, authentication failures, etc. However, obtaining reliable ground truth annotation of these anomalies in entire wireless networks is costly and time consuming. Under these circumstances, using data obtained through realistic network simulations is a common practice. 

In order to evaluate our model in the aforementioned scenario, we have followed the procedure of \cite{Anisa2017}, performing extensive network simulations in a typical Wi-Fi network setup (IEEE 802.11 WLANg 2.4 GHz in infrastructure mode) using OMNeT++~\cite{omnetpp} and INET~\cite{inet} simulators. Our network consists of 10 APs and 100 users accessing it. The pairwise distances between APs are known and fixed. Each sequence contains information about the traffic in a given AP during 10 consecutive hours and is divided in time slots of 15 minutes without overlap. Thus, every sequence has the same length, which is equal to 40 samples (time slots). Each sample contains the following 7 features: the number of unique users connected to the AP, the number of sessions within the AP, the total duration (in seconds) of association time of all current users, the number of octets transmitted and received in the AP and the number of packets transmitted and received in the AP. Anomalies typically occur for a limited amount of time within the whole sequence. However, in this experiment, we label a sequence as ``anomalous'' if there is at least one anomaly period in the sequence and we label it as ``normal'' otherwise. One of the simulations includes normal data only, while the remaining include both normal and anomalous sequences. In order to avoid contamination of normal data with anomalies that may occur simultaneously in other APs, we used the data of the normal simulation for training (150 sequences) and the remaining data for testing (378 normal and 42 anomalous sequences). 

In a Wi-Fi network, as users move in the covered area, they disconnect from one AP and they immediately connect to another in the vicinity. As such, the traffic in adjacent APs may be expected to be similar. Following this idea, the weight $\emG_{j,k}$, associated with the edge connecting nodes $j$ and $k$ in graph $\gG$, was set to the inverse distance between APs $j$ and $k$ and normalized so that $\max_{j,k} G_{j,k}=1$. As in \cite{Anisa2017}, sequences were preprocessed by subtracting the mean and dividing by the standard deviation and applying PCA, reducing the number of features to 3. For MHMM, we did 3-fold cross validation of the number of mixture components $M$ and hidden states per component $S$. We ended up using $M=15$ and $S=10$. We then used the same values of $M$ and $S$ for SpaMHMM and we did 3-fold cross validation for the regularization hyperparameter $\reg$ in the range $[10^{-4}, 1]$. The value $\reg=10^{-1}$ was chosen. We also cross-validated the number of hidden states in 1-HMM and K-HMM around the values indicated in \Secref{sec:experiments}. Every model was trained for 100 EM iterations or until the loss plateaus. For SpaMHMM, we did 100 iterations of the inner loop on each M-step, using a learning rate $\rho=10^{-3}$. 

Models were evaluated by computing the average log-likelihood per sample on normal and anomalous test data, plotting the receiver operating characteristic (ROC) curves and computing the respective areas under the curves (AUCs). \Figref{fig:roc} shows that the ROC curves for MHMM and SpaMHMM are very similar and that these models clearly outperform 1-HMM and K-HMM. This is confirmed by the AUC and log-likelihood results in \Tableref{tbl:wifi_results}. Although K-HMM achieved the best (lowest) average log-likelihood on anomalous data, this result is not relevant, since it also achieved the worst (lowest) average log-likelihood on normal data. This is in fact the model with the worst performance, as shown by its ROC and respective AUC.

The bad performance of K-HMM likely results mostly from the small amount of data that each of the $K$ models is trained with: in K-HMM, each HMM is trained with the data from the graph node (AP) that it is assigned to. The low log-likelihood value of the normal test data in this model confirms that the model does not generalize well to the test data and is probably highly biased towards the training data distribution. On the other hand, in 1-HMM there is a single HMM that is trained with the whole training set. However, the same HMM needs to capture the distribution of the data coming from all APs. Since each AP has its own typical usage profile, these data distributions are different and one single HMM may not be sufficiently expressive to learn all of them correctly. MHMM and SpaMHMM combine the advantages and avoid the disadvantages of both previous models. Clearly, since the mixtures for each node share the same dictionary of HMMs, every model in the mixture is trained with sequences from all graph nodes, at least in the first few training iterations. Thus, at this stage, the models may capture behaviors that are shared by all APs. As mixtures become sparser during training, some components in the dictionary may specialize on the distribution of a few APs. This avoids the problem observed in 1-HMM, which is unaware of the AP where a sequence comes from. We would also expect SpaMHMM to be sparser and have better performance than MHMM, but only the former supposition was true (see \Figref{fig:sparsity}). The absence of performance gains in SpaMHMM might be explained from the fact that this dataset consists of simulated data, where users are static (i.e. they do not swap between APs unless the AP where they are connected stops working) and so the assumption that closer APs have similar distributions does not bring any advantage.

\begin{table}
\begin{minipage}{.475\textwidth}
\centering
\includegraphics[width=0.8\linewidth]{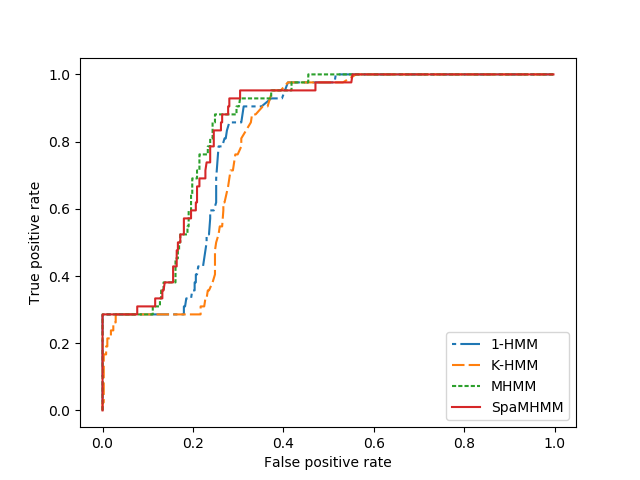}
\captionof{figure}{ROC curves for each model on the Wi-Fi dataset.}
\label{fig:roc}
\end{minipage}

\begin{minipage}{.475\textwidth}
\resizebox{\textwidth}{!}{
\begin{tabular}{l|c|c c}
\multicolumn{1}{c}{} & \multicolumn{1}{c}{} & \multicolumn{2}{c}{Avg. log-likelihood} \\
 & \multicolumn{1}{c|}{AUC} & Normal data & Anom. data \\
\hline
1-HMM & $0.806$ & $-6.28$ & $-112.75$ \\
K-HMM & $0.786$ & $-22.09$ & $\mathbf{-130.36}$ \\
MHMM & $\mathbf{0.842}$ & $-3.07$ & $-11.99$ \\
SpaMHMM & $0.839$ & $\mathbf{-3.06}$ & $-14.57$ \\
\end{tabular}}
\caption{AUC and average log-likelihood per sample for each model in
the Wi-Fi dataset. Best results are in
bold.}
\label{tbl:wifi_results}
\end{minipage}
\end{table}

\subsection{Human motion forecasting}
\label{sec:h36m}
The human body is constituted by several interdependent parts, which interact as a whole producing sensible global motion patterns. These patterns may correspond to multiple activities like walking, eating, etc. Here, we use our model to make short-time prediction of sequences of human joint positions, represented as motion capture (mocap) data. The current state of the art methodologies use architectures based on deep recurrent neural networks (RNNs), achieving remarkable results both in short-time prediction \cite{Fragkiadaki2015, Martinez2017} and in long-term motion generation \cite{Jain2016, Pavllo2018}.

Our experiments were conducted on the Human3.6M dataset from \cite{Ionescu2011, Ionescu2014}, which consists of mocap data from 7 subjects performing 15 distinct actions. In this experiment, we have considered only 4 of those actions, namely ``walking'', ``eating'', ``smoking'' and ``discussion''. There, the human skeleton is represented with 32 joints whose position is recorded at 50 Hz. We build our 32x32-dimensional symmetric matrix $\mG$ representing the graph $\gG$ in the following sensible manner: $G_{j,k}=1$, if there is an actual skeleton connection between joints $j$ and $k$ (e.g. the elbow joint is connected to the wrist joint by the forearm); $G_{j,k}=1$, if joints $j$ and $k$ are symmetric (e.g. left and right elbows); $G_{j,k}=0$, otherwise.

\subsubsection{Forecasting}

We reproduced as much as possible the experimental setup followed in \cite{Fragkiadaki2015}. Specifically, we down-sampled the data by a factor of 2 and transformed the raw 3-D angles into an exponential map representation. We removed joints with constant exponential map, yielding a dataset with 22 distinct joints, and pruned our matrix $\mG$ accordingly. Training was performed using data from 6 subjects, leaving one subject (denoted in the dataset by ``S5'') for testing. We did 3-fold cross-validation on the training data of the action ``walking'' to find the optimal number of mixture components $M$ and hidden states $S$ for the baseline mixture MHMM. Unsurprisingly, since this model can hardly overfit in such a complex task, we ended up with $M=18$ and $S=12$, which were the largest values in the ranges we defined. Larger values are likely to improve the results, but the training time would become too large to be practical. For SpaMHMM, we used these same values of $M$ and $S$ and we did 3-fold cross validation on the training data of the action ``walking'' to fine-tune the value of $\reg$ in the range $[10^{-4}, 1]$. We ended up using $\reg=0.05$. The number of hidden states in 1-HMM was set to 51 and in K-HMM it was set to 11 hidden states per HMM. The same values were then used to train the models for the remaining actions. Every model was trained for 100 iterations of EM  or until the loss plateaus. For SpaMHMM, we did 100 iterations of the inner loop on each M-step, using a learning rate $\rho=10^{-2}$.

In order to generate predictions for a joint (node) $y$ starting from a given prefix sequence $\mX_{\text{pref}}$, we build the distribution $p(\rmX | \mX_{\text{pref}}, y)$ (see details in \Secref{sec:posterior_proof}) and we sample sequences from that posterior. Our evaluation method and metric again followed \cite{Fragkiadaki2015}. We fed our model with 8 prefix subsequences with 50 frames each (corresponding to 2 seconds) for each joint from the test subject and we predicted the following 10 frames (corresponding to 400 miliseconds). Each prediction was built by sampling 100 sequences from the posterior and averaging. We then computed the average mean angle error for the 8 sequences at different time horizons. 

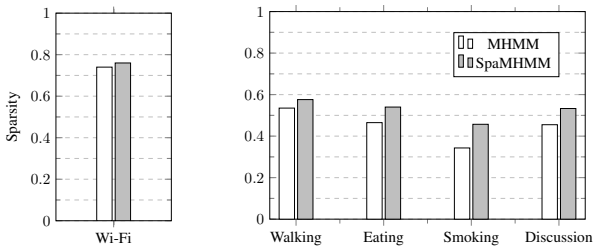
\begin{figure}
\centering
\resizebox{1.2\linewidth}{!}{
\begin{minipage}{.3\textwidth}
\begin{tikzpicture}
\begin{axis}[
	ybar,
    bar width=10pt,
    width=120pt,
    height=180pt,
    xmin=Wi-Fi, xmax=Wi-Fi,
	symbolic x coords={Wi-Fi},
    xtick=data,
    ylabel=Sparsity,
    ymin=0, ymax=1,
    ymajorgrids,
    major grid style={dashed},
    yminorgrids,
    minor grid style={dashed},
    minor tick num=1,
]
\addplot[fill=white, error bars/.cd, y dir=both, y explicit] 
	coordinates {(Wi-Fi, 0.74)};
\addplot[fill=lightgray, error bars/.cd, y dir=both, y explicit] 
	coordinates {(Wi-Fi, 0.76)};
\legend{}
\end{axis}
\end{tikzpicture}
\end{minipage}%
\begin{minipage}{.7\textwidth}
\begin{tikzpicture}
\begin{axis}[
	ybar,
    bar width=10pt,
    width=250pt,
    height=180pt,
	symbolic x coords={Walking,Eating,Smoking,Discussion},
    xtick=data,
    ymin=0, ymax=1,
    ymajorgrids,
    major grid style={dashed},
    yminorgrids,
    minor grid style={dashed},
    minor tick num=1,
    legend style={at={(0.75,0.9)},
	anchor=north,legend columns=1},
]
\addplot[fill=white] 
	coordinates {(Walking,0.535) (Eating,0.465) (Smoking,0.343) (Discussion,0.455)};
\addplot[fill=lightgray] 
	coordinates {(Walking,0.576) (Eating,0.540) (Smoking,0.457) (Discussion,0.533)};
\legend{MHMM,SpaMHMM}
\end{axis}
\end{tikzpicture}
\end{minipage}} 
\caption{Relative sparsity (number of coefficients equal to zero / total number of coefficients) of the obtained MHMM and SpaMHMM models on the Wi-Fi dataset (left) and on the Human3.6M dataset for different actions (right). Both models for the Wi-Fi dataset have 150 coefficients. All models for the Human3.6M dataset have 396 coefficients.}
\label{fig:sparsity}
\end{figure}

Results are in \Tableref{tbl:h36m_results}. Among our models (1-HMM, K-HMM, MHMM and SpaMHMM), SpaMHMM outperformed the remaining in all actions except ``eating''. For this action in particular, MHMM was slightly better than SpaMHMM, probably due to the lack of symmetry between the right and left sides of the body, which was one of the prior assumptions that we have used to build the graph $\gG$. ``Smoking'' and ``discussion'' activities may also be highly non-symmetric, but results in our and others' models show that these activities are generally harder to predict than ``walking'' and ``eating'. Thus, here, the skeleton structure information encoded in $\gG$ behaves as a useful prior for SpaMHMM, guiding it towards better solutions than MHMM. The worse results for 1-HMM and K-HMM likely result from the same limitations that we have pointed out in \Secref{sec:wi_fi}: each component in K-HMM is inherently trained with less data than the remaining models, while 1-HMM does not make distinction between different graph nodes. Extending the discussion to the state of the art solutions for this problem, we note that SpaMHMM compares favorably with ERD, LSTM-3LR and SRNN, which are all RNN-based architectures. Moreover, ERD and LSTM-3LR were designed specifically for this task, which is not the case for SpaMHMM. This is also true for GRU supervised and QuaterNet, which clearly outperform all remaining models, including ours. This is unsurprising, since RNNs are capable of modeling more complex dynamics than HMMs, due to their intrinsic non-linearity and continuous state representation. This also allows their usage for long-term motion generation, in which HMMs do not behave well due their linear dynamics and lack of long-term memory. However, unlike GRU supervised and QuaterNet, SpaMHMM  models the probability distribution of the data directly, allowing its application in domains like novelty detection. Regarding sparsity, the experiments confirm that the SpaMHMM mixture coefficients are actually sparser than those of MHMM, as shown in \Figref{fig:sparsity}.

\begin{table*}
\resizebox{\textwidth}{!}{
\begin{tabular}{l|c c c c| c c c c| c c c c | c c c c}
\multicolumn{1}{c}{} & \multicolumn{4}{c}{Walking} & \multicolumn{4}{c}{Eating} & \multicolumn{4}{c}{Smoking} & \multicolumn{4}{c}{Discussion} \\
miliseconds & 80 & 160 & 320 & 400 & 80 & 160 & 320 & 400 & 80 & 160 & 320 & 400 & 80 & 160 & 320 & 400 \\
\hline
1-HMM & 0.91 & 1.04 & 1.22 & 1.31 & 1.00 & 1.08 & 1.15 & 1.21 & 1.45 & 1.55 & 1.70 & 1.75 & 1.19 & 1.42 & 1.55 & 1.56 \\
K-HMM & 1.29 & 1.33 & 1.34 & 1.38 & 1.16 & 1.22 & 1.28 & 1.34 & 1.70 & 1.77 & 1.90 & 1.95 & 1.47 & 1.61 & 1.68 & 1.63 \\
MHMM & \tb{0.78} & \tb{0.93} & 1.13 & 1.21 & \tb{0.77} & \tb{0.87} & \tb{0.98} & \tb{1.06} & 1.44 & 1.53 & 1.69 & 1.77 & 1.14 & 1.36 & 1.52 & 1.54 \\
SpaMHMM & 0.80 & \tb{0.93} & \tb{1.11} & \tb{1.18} & 0.81 & 0.90 & 0.99 & \tb{1.06} & \tb{1.29} & \tb{1.39} & \tb{1.61} & \tb{1.67} & \tb{1.09} & \tb{1.30} & \tb{1.44} & \tb{1.49} \\
\hline
ERD \cite{Fragkiadaki2015} & 0.93 & 1.18 & 1.59 & 1.78 & 1.27 & 1.45 & 1.66 & 1.80 & 1.66 & 1.95 & 2.35 & 2.42 & 2.27 & 2.47 & 2.68 & 2.76 \\
LSTM-3LR \cite{Fragkiadaki2015} & 0.77 & 1.00 & 1.29 & 1.47 & 0.89 & 1.09 & 1.35 & 1.46 & 1.34 & 1.65 & 2.04 & 2.16 & 1.88 & 2.12 & 2.25 & 2.23 \\
SRNN \cite{Jain2016} & 0.81 & 0.94 & 1.16 & 1.30 & 0.97 & 1.14 & 1.35 & 1.46 & 1.45 & 1.68 & 1.94 & 2.08 & 1.22 & 1.49 & 1.83 & 1.93 \\
GRU sup. \cite{Martinez2017} & 0.28 & 0.49 & 0.72 & 0.81 & 0.23 & 0.39 & 0.62 & 0.76 & 0.33 & 0.61 & 1.05 & 1.15 & 0.31 & 0.68 & 1.01 & 1.09 \\
QuaterNet \cite{Pavllo2018} & \tu{0.21} & \tu{0.34} & \tu{0.56} & \tu{0.62} & \tu{0.20} & \tu{0.35} & \tu{0.58} & \tu{0.70} & \tu{0.25} & \tu{0.47} & \tu{0.93} & \tu{0.90} & \tu{0.26} & \tu{0.60} & \tu{0.85} & \tu{0.93}
\end{tabular}}
\caption{Mean angle error for short-term motion prediction on Human3.6M for different actions and time horizons. The results for ERD, LSTM-3LR, SRNN, GRU supervised and QuaterNet were extracted from \cite{Pavllo2018}. Best results among our models are in bold, best overall results are underlined.}
\label{tbl:h36m_results}
\end{table*}

\subsubsection{Joint cluster analysis}
\label{sec:spamhmm_cluster}
\begin{figure*}
\centering
\begin{minipage}{.5\textwidth}
  \centering
  \includegraphics[width=0.8\linewidth]{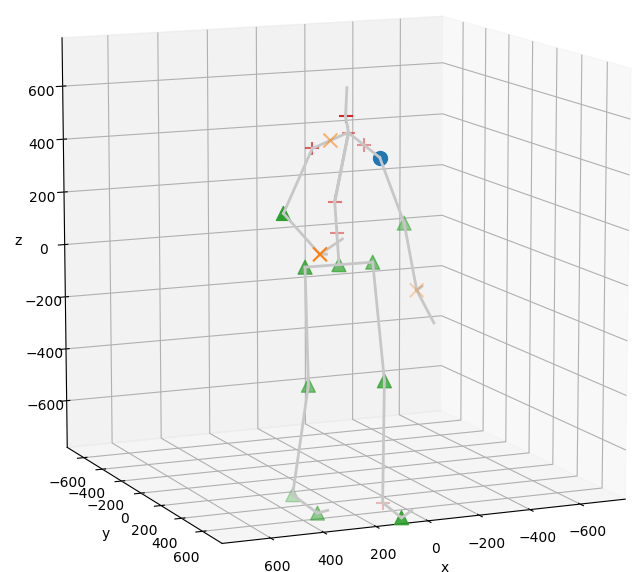}
\end{minipage}%
\begin{minipage}{.5\textwidth}
  \centering
  \includegraphics[width=0.8\linewidth]{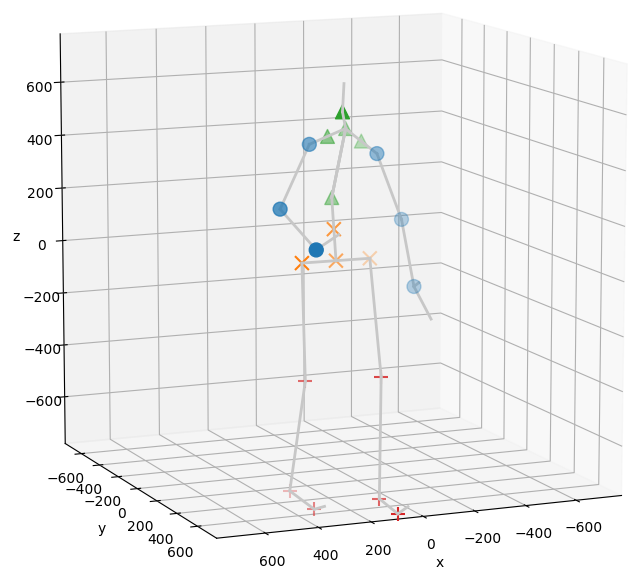}
\end{minipage}
\caption{Assignments of joints to clusters in MHMM (left) and SpaMHMM (right). The different symbols (`$\circ$', `$\smalltriangleup$', `x', `+') and the respective colors (blue, green, orange, red) on each joint represent the cluster that the joint was assigned to.}
\label{fig:clusters}
\end{figure*}

We may roughly divide the human body in four distinct parts: upper body (head, neck and shoulders), arms, torso and legs. Joints that belong to the same part naturally tend to have coherent motion, so we would expect them to be described by more or less the same components in our mixture models (MHMM and SpaMHMM). Since SpaMHMM is trained to exploit the known skeleton structure, this effect should be even more apparent in SpaMHMM than in MHMM. In order to confirm this conjecture, we have trained MHMM and SpaMHMM for the action ``walking'' using four mixture components only, i.e. $M=4$, and we have looked for the most likely component (cluster) for each joint:
\begin{equation}
C_k = \argmax_{m \in \{1,...,M\}} p(\rz=m | \ry=k) = \argmax_{m \in \{1,...,M\}} \alpha_{k,m},
\end{equation}
where $C_k$ is, therefore, the cluster assigned to joint $k$. The results are in \Figref{fig:clusters}. From there we can see that MHMM somehow succeeds on dividing the body in two main parts, by assigning the joints in the torso and in the upper body mostly to the red/'+' cluster, while those in the hips, legs and feet are almost all assigned to the green/'$\smalltriangleup$' cluster. Besides, we see that in the vast majority of the cases, symmetric joints are assigned to the same cluster. These observations confirm that we have chosen the graph $\gG$ for this problem in an appropriate manner. However, some assignments are unnatural: e.g. one of the joints in the left foot is assigned to the red/`+' cluster and the blue/`$\circ$' cluster is assigned to one single joint, in the left forearm. We also observe that the distribution of joints per clusters is highly uneven, being the green/`$\smalltriangleup$' cluster the most represented by far. SpaMHMM, on the other hand, succeeds on dividing the body in four meaningful regions: upper body and upper spine in the green/`$\smalltriangleup$' cluster; arms in the blue/`$\circ$' cluster; lower spine and hips in the orange/`x' cluster; legs and feet in the red/`+' cluster. Note that the graph $\gG$ used to regularize SpaMHMM does not include any information about the body part that a joint belongs to, but only about the joints that connect to it and that are symmetric to it. Nevertheless, the model is capable of using this information together with the training data in order to divide the skeleton in an intuitive and natural way. Moreover, the distribution of joints per cluster is much more even in this case, what may also help to explain why SpaMHMM outperforms MHMM: by splitting the joints more or less evenly by the different HMMs in the mixture, none of the HMM components is forced to learn too many motion patterns. In MHMM, we see that the green/`+' component, for instance, is the most responsible to model the motion of almost all joints in the legs and hips and also some joints in the arms and the red/`+' component is the prevalent on the prediction of the motion patterns of the neck and left foot, which are presumably very different.

\section{Conclusion and future work}
In this work we propose a method to model the generative distribution of sequential data coming from nodes connected in a graph with a known fixed topology. The method is based on a mixture over a shared dictionary of HMMs where the mixture coefficients are regularized during the learning process in such a way that affine nodes will tend to have similar coefficients, exploiting the known graph structure. We also prove that pairwise optimization of the coefficients leads to sparse mixtures. Experimental results suggest that sparsity holds in the general case. We evaluate the method's performance in two completely different tasks (anomaly detection in Wi-Fi networks and human motion forecasting), showing its effectiveness and versatility. 

For future work, we plan to extend/evaluate the usage of SpaMHMM for sequence clustering. This is an obvious extension that we did not explore thoroughly in this work, since its main focus was modeling the generative distribution of data. In this context, extending the idea behind SpaMHMM to mixtures of more powerful generative distributions is also in our plans. As is known, HMMs have limited expressiveness due to the strong independence assumptions they rely on. Thus, we plan to extend these ideas to develop an architecture based on more flexible generative models for sequence modeling, like those attained using deep recurrent architectures.

\section*{Acknowledgment}
The authors would like to thank Anisa Allahdadi for her great contribution on the development of the dataset for anomaly detection in Wi-Fi networks.



\bibliographystyle{IEEEtran}
\bibliography{IEEEabrv}
%
%
%

\appendix
\section{Appendix}
\label{sec:appendix}

\subsection{Derivation of \Algref{alg:mhmm}}
\label{sec:proof_em_noreg}
\Algref{alg:mhmm} follows straightforwardly from applying EM to the model defined by \twoeqrefs{hmm_mix}{hmm} with the objective \plaineqref{log_likelihood}. \\
Let us define the following notation: $\tX \coloneqq \left\lbrace \mX_i \right\rbrace_{i=1}^N$, $\vy \coloneqq \left\lbrace y_i \right\rbrace_{i=1}^N$, $\rvz \coloneqq \left\lbrace \rz_i \right\rbrace_{i=1}^N$ and $\rmH \coloneqq \left\lbrace \rvh_i \right\rbrace_{i=1}^N$. After building the usual variational lower bound for the log-likelihood and performing the E-step, we get the following well-known objective:
\begin{equation}
\label{m_step_obj_noreg}
\tilde{J}(\theta,\thetaprev) = \sum_{\rvz,\rmH} p(\tX,\rvz,\rmH | \vy,\thetaprev) \log p(\tX,\rvz,\rmH | \vy,\theta),
\end{equation}
which we want to maximize with respect to $\theta$ and where $\thetaprev$ are the model parameters that were kept fixed in the E-step. Some of the parameters in the model are constrained to represent valid probabilities, yielding the following Lagrangian:
\begin{align}
\label{lagrangian_noreg}
L(\theta,\thetaprev,\vlambda) = &\tilde{J}(\theta,\thetaprev) + \sum_k \lambda_k^{\text{mix}}\left(1 - ||\boldsymbol{\alpha}_k||_{1}\right) \nonumber\\
&+ \sum_{m,s} \lambda_{m,s}^\text{state} \left(1 - \sum_u A^m_{s,u}\right) \nonumber\\
&+ \sum_m \lambda_m^{\text{ini}} \left(1 - ||\boldsymbol{\pi}_m||_{1}\right),
\end{align}
where $\vlambda$ summarizes all Lagrange multipliers used here (not to be confused with the regularization hyperparameter $\reg$ used in \eqref{objective}). Differentiating \eqref{lagrangian_noreg} with respect to each model parameter and Lagrange multiplier and solving for the critical points yields:

{\scriptsize
\begin{align}
	& \evalpha_{k,m} = \frac{\sum_i p(\rz_i=m |\mX_i,y_i,\thetaprev) \Indyk}{\sum_i \Indyk},\\
    & \evpi_{m,s} = \frac{\sum_i p(\rz_i=m | \mX_i,y_i,\thetaprev) p(\rh_i^{(0)}=s | \rz_i=m,\mX_i,y_i,\thetaprev)}{\sum_i p(\rz_i=m | \rmX_i,y_i,\thetaprev)}, \\
    \lefteqn{\emA^m_{s,u} =} \nonumber\\
    & \frac{\sum_i p(\rz_i=m | \mX_i,y_i,\thetaprev) \sum_{t=1}^{T_i} p(\rh_i^{(t-1)}=s, \rh_i^{(t)}=u | \rz_i=m,\mX_i,y_i,\thetaprev)} {\sum_i p(\rz_i=m | \mX_i,y_i,\thetaprev) \sum_{t=1}^{T_i} p(\rh_i^{(t-1)}=s | \rz_i=m,\mX_i,y_i,\thetaprev)}, \\
    & \vmu_{m,s} = \frac{\sum_i p(\rz_i=m | \mX_i,y_i,\thetaprev) \sum_{t=1}^{T_i} p(\rh_i^{(t)}=s | \rz_i=m,\mX_i,y_i,\thetaprev) \vx_i^{(t)}}{\sum_i p(\rz_i=m | \mX_i,y_i,\thetaprev) \sum_{t=1}^{T_i} p(\rh_i^{(t)}=s | \rz_i=m,\mX_i,y_i,\thetaprev)}, \\
    \lefteqn{\vsigma^2_{m,s} =} \nonumber\\
    & \frac{\sum_i p(\rz_i=m | \mX_i,y_i,\thetaprev) \sum_{t=1}^{T_i} p(\rh_i^{(t)}=s | \rz_i=m,\mX_i,y_i,\thetaprev) \left(\vx_i^{(t)} - \boldsymbol{\mu}^m_s\right)^2}{\sum_i p(\rz_i=m | \mX_i,y_i,\thetaprev) \sum_{t=1}^{T_i} p(\rh_i^{(t)}=s | \rz_i=m,\mX_i,y_i,\thetaprev)}, \\
    & \forall \, k,m,s,u. \nonumber
\end{align}}%
Defining $n_k$, $\rho_{i,m}$, $\gamma_{i,m,s}$ and $\xi_{i,m,s,u}$ as in \Algref{alg:mhmm} the result follows.

\subsection{Derivation of \Algref{alg:spamhmm}}
\label{sec:proof_em_reg}
Using the same notation as in \Secref{sec:proof_em_noreg}, we may rewrite \eqref{objective} as:
\begin{align}
J_r(\theta) = &\frac{1}{N}\log \sum_{\rvz,\rmH} p(\tX,\rvz,\rmH | \vy,\theta) \nonumber\\
& +\frac{\reg}{2} \sum_{j,k\neq j} \emG_{j,k} \E_{\rz \sim p(\rz | \ry=j, \theta)} [ p(\rz | \ry=k, \theta) ].
\end{align}
Despite the regularization term, we may still lower bound this objective by introducing a variational distribution $q(\rvz,\rmH)$ and using Jensen's inequality in the usual way:
\begin{align}
J_r(\theta) &\geq \frac{1}{N} \E_{\rvz,\rmH \sim q} \left[\log \frac{p(\tX,\rvz,\rmH | \vy,\theta)}{q(\rvz,\rmH)} \right] \nonumber\\
&\mathbin{\hphantom{=}}{} + \frac{\reg}{2} \sum_{j,k\neq j} \emG_{j,k} \E_{\rz \sim p(\rz | \ry=j, \theta)} [ p(\rz | \ry=k, \theta) ] \nonumber\\
&\coloneqq V_r(\theta, q).
\end{align}
Clearly,
\begin{align}
\lefteqn{J_r(\theta) - V_r(\theta, q) =}\nonumber\\
&=\frac{1}{N} \left(\log p(\tX|\vy,\theta) - \E_{\rvz,\rmH \sim q} \left[\log \frac{p(\tX,\rvz,\rmH | \vy,\theta)}{q(\rvz,\rmH)} \right] \right) \nonumber\\
&= \frac{1}{N} \KL \left(q(\rvz,\rmH) || p(\rvz,\rmH | \tX,\vy,\theta)  \right),
\end{align}
which, fixing the parameters $\theta$ to some value $\thetaprev$ and minimizing with respect to $q$, yields the usual solution $q^*(\rvz,\rmH) = p(\rvz,\rmH | \tX,\vy,\thetaprev)$. Thus, in the M-step, we want to find:
\begin{align}
\lefteqn{\argmax_\theta V_r(\theta, q^*) =} \nonumber\\
& =\argmax_\theta \frac{1}{N}\sum_{\rvz,\rmH} p(\tX,\rvz,\rmH | \vy,\thetaprev) \log p(\tX,\rvz,\rmH | \vy,\theta) \nonumber\\
& \mathbin{\hphantom{=}}{}+ \frac{\reg}{2} p(\tX|\vy,\thetaprev) \sum_{j,k\neq j} \emG_{j,k} \E_{\rz \sim p(\rz | \ry=j, \theta)} [ p(\rz | \ry=k, \theta) ] \nonumber\\
& = \argmax_\theta \frac{1}{N}\tilde{J}(\theta, \thetaprev) + \reg R(\theta, \thetaprev) \nonumber\\
& \coloneqq \argmax_\theta \tilde{J}_r(\theta, \thetaprev),
\end{align}
where $\tilde{J}(\theta, \thetaprev)$ is as defined in \eqref{m_step_obj_noreg} and $R(\theta, \thetaprev)$ is our regularization (weighted by the data likelihood), which is simply a function of the parameters $\valpha_1,...,\valpha_K$:
\begin{align}
R(\theta, \thetaprev) &= \frac{1}{2}p(\tX|\vy,\thetaprev) \sum_{j,k\neq j} \emG_{j,k} \E_{\rz \sim p(\rz | \ry=j, \theta)} [ p(\rz | \ry=k, \theta) ] \nonumber\\
&= \frac{1}{2}p(\tX|\vy,\thetaprev) \sum_{j,k\neq j} \emG_{j,k} \valpha_j \transp \valpha_k \nonumber\\
&= R(\valpha_1,...,\valpha_K,\thetaprev).
\end{align}
Now, we may build the Lagrangian as done in \Secref{sec:proof_em_noreg}. Since $R$ only depends on the $\valpha$'s, the update equations for the remaining parameters are unchanged. However, for the $\valpha$'s, it is not possible to obtain a closed form update equation. Thus, we use the reparameterization defined in \eqref{normalization} and update the new unconstrained parameters $\vbeta$ via gradient ascent. \\
We have:
\begin{align}
& \frac{\partial \tilde{J}}{\partial \alpha_{k,m}} = \frac{p(\tX|\vy,\thetaprev)}{\alpha_{k,m}} \sum_i p(\rz_i=m | \mX_i,y_i,\thetaprev) \Indyk \label{dJ_dalpha}, \\
& \frac{\partial R}{\partial \alpha_{k,m}} = p(\tX|\vy,\thetaprev) \sum_{j \neq k} \emG_{j,k} \alpha_{j,m}. \label{dR_dalpha}
\end{align}
From \twoeqrefs{dJ_dalpha}{dR_dalpha}, we see that the the resulting gradient $\nabla_{\boldsymbol{\alpha}_k} \tilde{J}_r = \frac{1}{N}\nabla_{\boldsymbol{\alpha}_k} \tilde{J} + \reg \nabla_{\boldsymbol{\alpha}_k} R$ is equal to some vector scaled by the joint data likelihood $p(\tX|\vy,\thetaprev)$, which we discard since it only affects the learning rate, besides being usually very small and somewhat costly to compute. This option is equivalent to using a learning rate that changes at each iteration of the outter loop of the algorithm. \\
\Eqref{normalization} yields the following derivatives:
\begin{align}
& \frac{\partial \alpha_{k,m}}{\partial \beta_{k,m}} = \Indbeta \frac{2\sigma'(\beta_{k,m})}{\sigma(\beta_{k,m})} \alpha_{k,m} (1 - \alpha_{k,m}), \\
& \frac{\partial \alpha_{k,m}}{\partial \beta_{k,l}} = \Indbeta \frac{-2\sigma'(\beta_{k,m})}{\sigma(\beta_{k,m})} \alpha_{k,m} \alpha_{k,l}, \text{ for } l \neq m.
\end{align}
Finally, by the chain rule, we obtain:
\begin{align}
\lefteqn{\frac{\partial \tilde{J}}{\partial \beta_{k,m}} = \sum_l \frac{\partial \tilde{J}}{\partial \alpha_{k,l}} \frac{\partial \alpha_{k,l}}{\partial \beta_{k,m}}} \nonumber\\
&= \Indbeta \frac{2\sigma'(\beta_{k,m})}{\sigma(\beta_{k,m})} \sum_i  \left(p(\rz_i=m | \mX_i,y_i,\thetaprev) -\alpha_{k,m}\right)\Indyk, \\
\lefteqn{\frac{\partial R}{\partial \beta_{k,m}} = \sum_l \frac{\partial R}{\partial \alpha_{k,l}} \frac{\partial \alpha_{k,l}}{\partial \beta_{k,m}}} \nonumber\\
&= \Indbeta \frac{2\sigma'(\beta_{k,m})}{\sigma(\beta_{k,m})} \alpha_{k,m}\sum_{j \neq k} \emG_{j,k}\left(\alpha_{j,m} - \valpha_j \transp \valpha_k \right).
\end{align}
Defining $\delta_{k,m} \coloneqq \frac{\partial \tilde{J}_r}{\partial \beta_{k,m}} = \frac{1}{N}\frac{\partial \tilde{J}}{\partial \beta_{k,m}} + \reg \frac{\partial R}{\partial \beta_{k,m}}$ and applying the gradient ascent update formula to $\beta_{k,m}$ the result follows.

\subsection{Getting the posterior distribution of observations in SpaMHMM}
\label{sec:posterior_proof}
In this section, we show how to obtain the posterior distribution $p(\rmX | \mX_\text{pref}, y)$ of sequences $\rmX = \left(\rvx^{(1)}, ...,\rvx^{(T)}\right)$ given an observed prefix sequence $\mX_\text{pref} = \left(\vx^{(-T_\text{pref}+1)}, ...,\vx^{(0)}\right)$, both coming from the graph node $y$. We consider the case where $p(\rmX | \ry)$ is a SpaMHMM (or MHMM) model and so, using \eqref{hmm_mix}, we have:
\begin{align}
\label{mhmm_posterior}
p(\rmX | \mX_\text{pref}, y)&= \sum_{\rz} p(\rmX | \rz, \mX_\text{pref}, y) p(\rz | \mX_\text{pref}, y) \nonumber\\
&= \sum_{\rz} p(\rmX | \rz, \mX_\text{pref}) p(\rz | \mX_\text{pref}, y),
\end{align}
where the second equality follows from the fact that the observations $\rmX$ are independent from the graph node $\ry$ given the latent variable $\rz$. The posterior $p(\rz | \mX_\text{pref}, y)$ may be obtained as done in \Algref{alg:mhmm}, so we now focus on $p(\rmX | \rz, \mX_\text{pref})$, which follows from \eqref{hmm}:
\begin{align}
\label{component_posterior}
\lefteqn{p(\rmX | \rz, \mX_\text{pref}) =} \nonumber\\
&=\sum_{\rvh} p(\rh^{(0)} | \rz, \mX_\text{pref}) \prod_t \Biggl( p(\rh^{(t)} | \rh^{(t-1)}, \rz, \mX_\text{pref}) \nonumber\\
&\hphantom{=\sum_{\rvh} p(\rh^{(0)} | \rz, \mX_\text{pref}) \prod_t \Biggl(~} p(\rvx^{(t)} | \rh^{(t)}, \rz, \mX_\text{pref})\Biggr) \nonumber\\
&= \sum_{\rvh} p(\rh^{(0)} | \rz, \mX_\text{pref}) \prod_t p(\rh^{(t)} | \rh^{(t-1)}, \rz) p(\rvx^{(t)} | \rh^{(t)}, \rz),\nonumber\\
\end{align}
where we have used the independence assumptions of the HMM. Here, the initial state posteriors $p(\rh^{(0)} | \rz, \mX_\text{pref})$ are actually the final state posteriors for the sequence $\mX_\text{pref}$ for each HMM in the mixture, so they can also be computed as indicated \Algref{alg:mhmm}.

Thus, we see that, in order to obtain the posterior $p(\rmX | \mX_\text{pref}, y)$, we only need to update the mixture coefficients $p(\rz | \mX_\text{pref}, y)$ and the initial state probabilities $p(\rh^{(0)} | \rz, \mX_\text{pref})$. All remaining parameters are unchanged.
\end{document}